\documentclass{article}

     \usepackage[final]{neurips_2019}

\usepackage[utf8]{inputenc} 
\usepackage[T1]{fontenc}    
\usepackage{hyperref}       
\usepackage{url}            
\usepackage{booktabs}       
\usepackage{amsfonts}       
\usepackage{nicefrac}       
\usepackage{microtype}      
\usepackage{bm}
\usepackage{graphicx}
\usepackage{multirow}
\usepackage{natbib}
\newcommand{\MDP}{\mathcal{M}}
\newcommand{\Obs}{\mathcal{S}}
\newcommand{\Act}{\mathcal{A}}
\newcommand{\Rew}{r}
\newcommand{\Trans}{T}
\newcommand{\LongTermRew}{R}

\newcommand{\Reals}{\mathbb{R}}
\newcommand{\Naturals}{\mathbb{N}}

\newcommand{\param}{\bm{\theta}}
\newcommand{\grad}{\bm{g}}
\newcommand{\MaxTasks}{N_{max}}

\title{Reinforcement Learning of\\ Multi-Domain Dialog Policies Via Action Embeddings}

\author{%
  Jorge A. Mendez\thanks{Work done while the author was at Facebook AI.} \\
  University of Pennsylvania \\
  \texttt{mendezme@seas.upenn.edu} \\
  \And
  Alborz Geramifard\\
  Facebook AI\\
  \texttt{alborzg@fb.com} \\
  \AND
  Mohammad Ghavamzadeh\\
  Facebook AI Research\\
  \texttt{mgh@fb.com} \\
  \And
  Bing Liu\\
  Facebook Assistant\\
  \texttt{bingl@fb.com} \\
}

\begin{document}

\maketitle

\begin{abstract}
Learning task-oriented dialog policies via reinforcement learning typically requires large amounts of interaction with users, which in practice renders such methods unusable for real-world applications. In order to reduce the data requirements, we propose to leverage data from across different dialog domains, thereby reducing the amount of data required from each given domain. In particular, we propose to learn domain-agnostic \textit{action embeddings}, which capture general-purpose structure that informs the system how to act given the current dialog context, and are then specialized to a specific domain. We show how this approach is capable of learning with significantly less interaction with users, with a reduction of 35\% in the number of dialogs required to learn, and to a higher level of proficiency than training separate policies for each domain on a set of simulated domains.
\end{abstract}

\section{Introduction}

Conversational systems provide a seamless interaction mechanism for users, allowing them to explore the capabilities of the system without the need to learn specialized instructions. In particular, task-oriented dialog systems have the goal of helping users achieve their goals in a particular domain. For example, a user may want to find a restaurant of a particular cuisine in a certain area of town, or figure out which bus to take to get from one place to another. Traditionally, these tasks have been modelled as slot-filling problems, and dialog policies have been hand-crafted for the system to respond to each foreseeable conversation context. In recent years, researchers have focused on using machine learning (ML) techniques for automatically learning such policies~\citep{Gao2019Neural}. Although ML has been shown to substantially improve the quality of dialog systems, it comes at the cost of needing vast amounts of data, which has unfortunately prevented its widespread adoption by industry practitioners. 

As a solution to this problem, several recent efforts have leveraged data from multiple dialog domains in order to reduce the amount of data required for each specific domain. A large majority of these approaches have focused on learning to imitate policies from ground-truth actions observed in a dialog corpus using supervised learning techniques. The downside of this type of imitation learning is that the quality of the learned policy is bounded by the quality of the policy used to collect the data, which is often suboptimal. On the other hand, learning the policy via reinforcement learning (RL) would allow the agent to surpass the quality of the policy used to generate the corpus.

We propose a multi-domain learning approach to learning dialog policies via RL. The core idea of our approach is to automatically learn a set of action embeddings that can be re-used across all the various domains. Ideally, such embeddings can help accelerate the learning process of the RL agent by discovering some hidden structure common to all domains. Intuitively, this structure can indicate simple relations like the need to find out the values for all slots in a slot-filling task, or more complex ones like how much noise to tolerate before requesting confirmation for a slot value. 

Our approach is fully neural-based, where each component of the dialog system is represented by a neural network. Each component is trained separately, in order to avoid interference between the modules, but the proposed architecture could be trained end-to-end without modifications. We evaluate our method in two multi-domain settings: batch multi-task learning and transfer learning. We show that our method is capable of learning more quickly than the single-domain approach, requiring 35\% less interaction with users on average, and also achieves higher final performance.

\section{Related work}

Recent years have seen a steep increase in the amount of work on applying ML techniques to the problem of conversational systems. In this section, we review connections to work closest to ours. We focus on approaches for tackling task-oriented conversational systems and omit those addressing open-domain agents, as they require considerably different methods. We refer readers to relevant surveys for a comparison of methods from the two settings~\citep{Gao2019Neural, chen2017survey}.

The primary approach to leveraging data for learning task-oriented dialog policies is to directly mimic policies from dialog corpora~\citep{shu2018incorporating, williams2017hybrid, bordes2017learning,lipton2018bbqnetworks}. In this setting, the system operates under the assumption that the policy used for data collection is optimal for (or at least proficient at) the task at hand, and treats the actions in the corpus as ground-truth labels for the action to take given the context of the dialog. Then, the problem reduces to selecting how to model the state of the conversation and the connection between the state and the action. Most recent methods use deep neural network architectures for these models~\citep{Gao2019Neural,bordes2017learning,qian2019domain,zhao2018zeroshot}. The key idea is that each module in a conversational system can be represented by a neural network, and the overall model can be constructed by stacking the modules one after the other. Training all models jointly introduces potential interference between the different modules and typically requires larger amounts of data, but it is possible to train models with less supervision, as intermediate labels are not required. On the contrary, training them separately can be done with less data, but the data must be properly annotated with labels for each module. In this work, we focus on the latter technique because our focus is on reducing the data requirements, but our proposed architecture could also be trained via the former. 

The main disadvantage of training  the system via supervised labels is the assumption of optimality of the underlying policy. Specifically, supervised learning assumes access to the \textit{correct} action, and so it is not possible for the system to learn a policy better than the observed one. One alternative is for the system to autonomously find the optimal policy via RL, explicitly training to maximize its proficiency regardless of the quality of the policy in the given data set~\citep{zhao2016towards, fazel2017learning, zhou2017endto, lipton2018bbqnetworks, kim2014use}. Most current RL algorithms are applicable to models with the architecture described for supervised learning, so the design of the system can leverage work from the supervised setting. In fact, it is possible to initialize the policy for RL training via supervised learning~\citep{liu2017endto}. The biggest hurdle to overcome in training such systems via RL is the requirement for vast amounts of interaction with users. 

A recently proposed solution to reduce the amount of data required to train a dialog system is to leverage data from across different dialog domains. In the simplest setting, a multi-domain architecture can be constructed to track the state of the dialog, which has typically been achieved by training neural networks with certain parameters shared across all domains and certain domain-specific parameters~\citep{shah2019robust, liu2017multidomain, rastogi2017scalable}. Once the state of the dialog is tracked, domain-specific policies can be trained on the resulting state. However, note that this method only allows the system to leverage the multi-domain structure for finding the state representation, but fails to exploit commonalities between the ways to act in different domains.

A better approach is to learn a dialog agent's policy for multiple domains~\citep{vlasov2018fewshot}, potentially in an end-to-end fashion~\citep{qian2019domain, zhao2018zeroshot, lee2017toward}. However, most work for learning such multi-domain policies has been focused on training supervised models, under the assumption that the given policy is optimal. 

Work on multi-domain RL for dialog policies is substantially more sparse. A majority of efforts in this setting have either used shallow learning methods, which are not capable of handling arbitrary state representations~\citep{gasic2013pomdp, gavsic2017dialogue, wang2015learning}, or have focused on the setting where different domains share a significant subset of the slots and actions, and so parts of the policy can be directly shared across domains~\citep{chen2018policy}. All these methods heavily rely on hand-specified connections across domains. On the contrary, our proposed solution is designed to handle arbitrary domains, with no \textit{a priori} knowledge of the connections between them. These connections are automatically found by our algorithm in the form of hidden \textit{action embeddings}, vectors that capture a domain-agnostic representation of the optimal action to take at the current state of the dialog. 

Another related, but distinct, problem that has received attention in recent years is that of learning to solve composite dialog tasks~\citep{cuayahuitl2017scaling, cuayhuitl2016deep, peng2017composite, wang2014policy, budzianowski2017sub, tang2018subgoal}. In this setting, the agent's goal is to learn to distinguish between different domains and select a domain-specific policy accordingly. However, they do not improve one domain's performance based on information from  another domain as we do.

Our work is also tangentially related to prior efforts on using action embeddings for RL~\citep{he2016deep,dulac2015deep}. In this line, the agent is given a description of the action (e.g., in natural language), and embeds it into a low-dimensional space. In contrast, our work determines the action embeddings autonomously, based solely on interactions with users in different domains. 

\section{The multi-domain dialog policy learning problem}

We begin by formalizing dialog policy learning as an RL problem. Then, we extend our definition to the multi-domain setting and explicitly state two multi-domain problem variants. 

\subsection{Dialog policy learning via RL}
\label{sec:DialogPolicyLearning}

The first step towards solving any problem via RL is constructing a Markov decision process (MDP), given by a tuple $\MDP = \langle \Obs, \Act, \Rew, \Trans, \gamma \rangle$. $\Obs \subseteq \Reals^n$ is the observation space, the set of inputs based upon which the agent can make decisions. In dialog learning, these observations should capture the entire context of the conversation in order to satisfy the Markov assumption. $\Act \subseteq \Naturals$ is the set of actions the system can execute. These actions could be, for example, dialog acts, like querying the user for the type of food he or she is looking for. $\Rew : \Obs \mapsto \Reals$ is the reward function that indicates how good or bad a particular state is. A prototypical example of a reward for dialog policy learning is a small negative reward for every step and a large positive reward for the final state in a successful execution of the task. $\Trans: \Obs \times \Act \times \Obs \mapsto [0, 1]$ is the transition probability function, which indicates the likelihood of encountering a given observation after executing an action from a given previous observation. In a conversational setting, this transition function is highly stochastic, and is determined by the user. 

Given an MDP, the goal of the agent is to learn a policy $\pi_{\param}$ that optimizes the long term rewards ${\LongTermRew = \sum_{t=0}^{H} \gamma ^{t} \Rew(s_t)}$. The discount factor $\gamma$ of the MDP prescribes a trade-off between immediate and future rewards. The policy $\pi_{\param}: \Obs \times \Act \mapsto [0,1]$, parameterized by a vector $\param$, stochastically dictates the agent's behavior, by specifying the probability of taking each action at any given state. As such, the objective of our learning problem reduces to finding the optimal set of parameters $\param$ to yield a policy that maximizes the long-term rewards $\LongTermRew$\footnote{Note that we are considering parametric policy learning, but we could similarly parameterize a value function in order to use value-based learning algorithms.}. 

\subsection{Multi-domain dialog policy learning}

Based on our definition for dialog policy learning, we frame our multi-domain problem as multi-task RL. In particular, we assume our agent will face a set of tasks $\{\MDP^{(1)},\MDP^{(2)},\ldots,\MDP^{(\MaxTasks)}\}$, each corresponding to a different dialog domain: $\MDP^{(t)} = \langle \Obs, \Act^{(t)}, \Rew^{(t)}, \Trans^{(t)}, \gamma \rangle$. We deliberately make the actions, rewards, and transitions domain-dependent, but insist that the observation space be shared across domains. This turns out to be crucial for our architecture design, as we will see in Section~\ref{sec:MultiDomainTransfer}, but we note that it does not impose restrictions on the domains, but rather on the architecture of the state tracking module. The goal of the agent is then to find a set of parameters $\param$ that yields a set of policies $\{\pi_{\param}^{(1)}, \pi_{\param}^{(2)}, \ldots, \pi_{\param}^{(\MaxTasks)} \}$ that perform well on their corresponding domains. 

This definition allows us to consider various modalities of multi-domain learning. We review the two that we will evaluate in our experiments, batch multi-task and transfer learning, but our framework is more general and can also accommodate other modalities, such as continual or lifelong learning.

\paragraph{Batch multi-task learning.} In the simplest setting of batch \textit{multi-task learning} (MTL), we assume that the agent encounters all domains simultaneously, and its goal is to optimize the average performance across all of them. The learning objective is then to maximize the average long-term rewards $\frac{1}{\MaxTasks} \sum_{t=1}^{\MaxTasks} \LongTermRew^{(t)}$. This problem setting corresponds to a frequently encountered situation in which a new conversational system for handling multiple domains is being designed, and the creators wish to leverage limited data available from each of the domains.

\paragraph{Transfer learning.} Similarly, we consider what happens if the system designers now want to add a new domain to their previously deployed system. Ideally, they would be able to leverage any structure learned from the initial set of domains in order to quickly learn the new one. This is the problem addressed by \textit{transfer learning} (TL), where the data from a set of source domains is used in order to improve the performance on a target domain with limited data\footnote{Recent literature refers to a setting closely matching this one with the term \textit{meta-learning}. Even though subtle distinctions separate the two, we consider them a single framework for the purposes of this work.}. In this situation, the objective becomes to optimize the performance on the new domain $\LongTermRew^{(\MaxTasks+1)}$ with the limited amount of data available, by exploiting the structure learned from training on tasks $1, 2, \ldots, \MaxTasks$.

\section{Multi-domain transfer via action embeddings}
\label{sec:MultiDomainTransfer}
We now present our model architecture for training the multi-domain policies, and then detail the training procedures used for the two problem settings: MTL and TL.

\begin{figure}[t]
    \centering
    \includegraphics[width=0.8\textwidth]{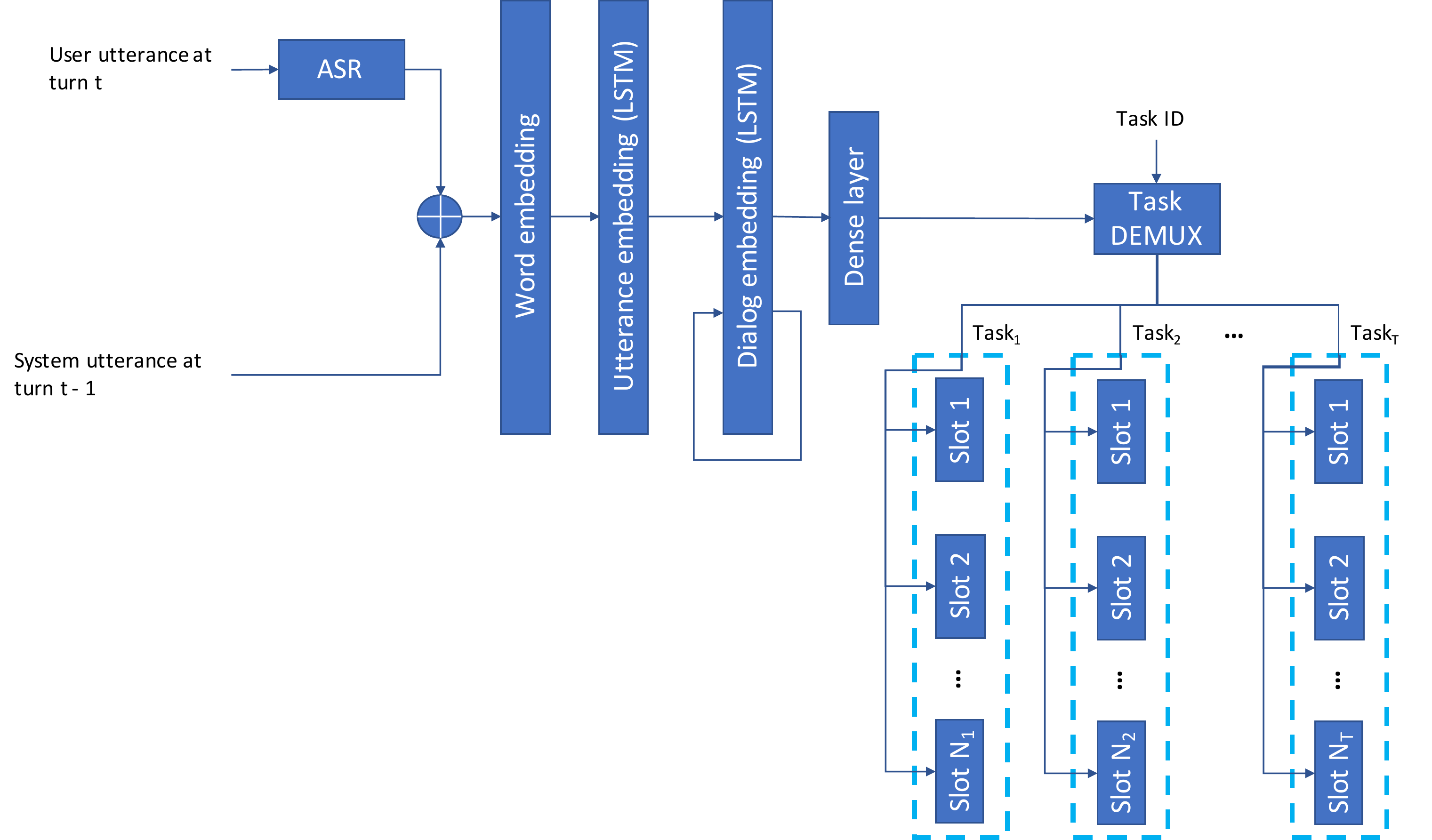}
    \caption{Multi-domain dialog state tracker. Word, utterance, and dialog embedding layers, as well as an additional fully connected layer, are shared across all domains. A separate classification head is used for each domain, separated in the figure by dashed lines. $\oplus$ indicates concatenation.}
    \label{fig:MultiDomainDST}
\end{figure}

\subsection{Multi-domain policy architecture}
\label{sec:MultiDomainArchitecture}
The first step in constructing the MDP for each domain is crafting the observation space. For our multi-domain problem, we have two requirements: 1) the observations must capture the current state of the conversation, and 2) the space must be common to all domains. We achieve this by training a neural dialog state tracker (DST) as shown in Figure~\ref{fig:MultiDomainDST}. In summary, the concatenated text from the user's utterance and the previous system's utterance are used as inputs, which are then passed through consecutive word, utterance, and dialog embedding layers. The dialog embedding layer synthesizes all the information from the dialog history at any given point, and thus is used as the observation for the RL module. In order to train this system, we add an additional hidden layer common to all domains, and add separate classification heads for the slot values of each of the domains. Training is done in a supervised MTL fashion using the average cross-entropy loss across domains and slots, selecting the relevant head based on ground-truth task indicators.

\begin{figure}[t]
    \centering
    \includegraphics[width=0.8\textwidth]{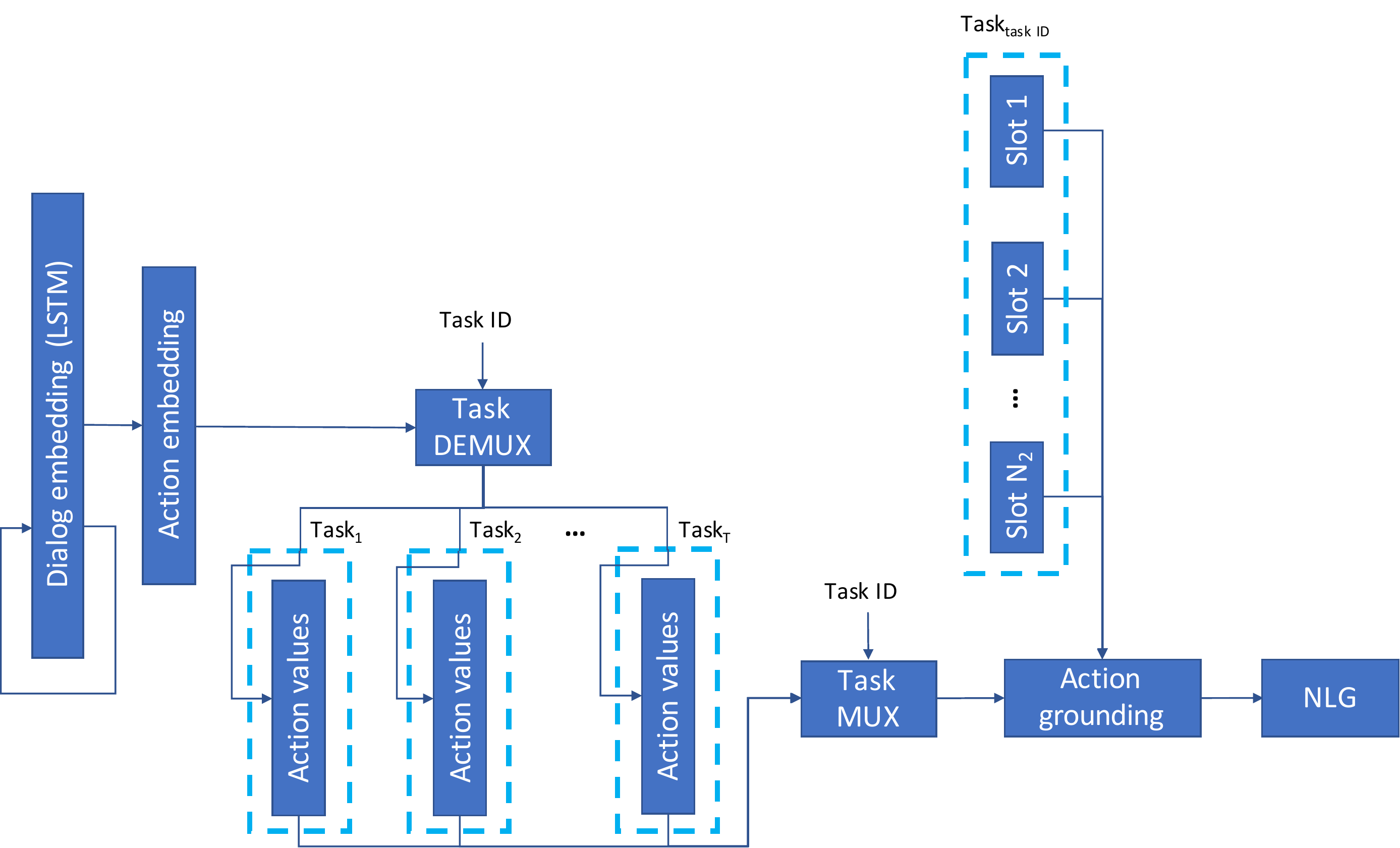}
    \caption{Multi-domain policy model. The input space is common to all domains. An action embedding layer is trained to detect commonalities in how to act for each domain. Separate actions heads are used for each domain, and selected actions are then grounded on the specific slot values.}
    \label{fig:MultiDomainRL}
\end{figure}

Given a trained DST module, we design our policy model as presented in Figure~\ref{fig:MultiDomainRL}. As mentioned before, the input to the policy module is the dialog embedding layer from the DST. The input is passed through an \textit{action embedding} layer. This layer, shared across all domains, is the central piece of our model. It autonomously learns the relations between the different policies required to solve the various tasks. Crucially, this is substantially different from the majority of approaches to multi-domain dialog learning, which limit the sharing of information to state tracking and disregard the potential benefits of sharing information about how to act. In our experiments, the action embedding layer is a simple fully connected layer, trained jointly with the rest of the network. Based on the task indicator, the relevant action selection head is picked. Given a chosen action, the dialog acts are grounded based on the DST output and passed through an off-the-shelf natural language generator (NLG).

\subsection{Multi-domain policy training via RL}
\label{MultiDomainTraining}
The specific method for training the model above varies depending on the exact problem setting. However, for all of them, it is possible to select a base learning algorithm that is wrapped around by a higher level multi-domain method. We use trust region policy optimization (TRPO)~\citep{schulman2015trust} for our experiments, as it is one of the strongest and most stable policy learning methods available. Note that it would be possible to similarly train our model with any other policy learning algorithm. It would also be possible to do so with value-based learning approaches (e.g., deep Q-networks~\citep{mnih2015human}), but these do not incorporate exploration explicitly in the policy model, which is necessary for accelerating learning based on parameter sharing. 

More concretely, assume that we use a policy gradient algorithm whose gradient estimate for the cost function of task $t$ is $\grad^{(t)}$. With this, in the MTL case, the aggregate gradient for all tasks is $\frac{1}{\MaxTasks} \sum_{t=1}^{\MaxTasks} \grad^{(t)}$. In practice, this means that, for every training iteration, we must obtain data for all $\MaxTasks$ tasks and compute their gradients, and then average them in order to obtain the gradient for the aggregate cost. In the TL case, a simple solution is to use the MTL formulation to train a network that works well on the first $\MaxTasks$ tasks, and then use the resulting network as initialization for the $(\MaxTasks+1)-$th task. Note that, in both cases, since the final layer is completely domain-specific, the action selection portion of the network for each task is trained only on the gradient for that task. More critically, in the TL case, the initial network only contains some \textit{hidden} representation, in the form of action embeddings, for solving the task, but the initial policy itself is completely random.

\section{Experimental results}
We evaluated our proposed algorithm using the simulated data set SimDial~\citep{zhao2018zeroshot}. The data set, originally constructed for end-to-end dialog generation, consists of six task-oriented dialog domains: bus, movie, weather, restaurant, restaurant-slot, and restaurant-style. Restaurant-slot differs from restaurant in that the slot values are different, and restaurant-style uses different NLG templates for generating the dialogs. In our experiments, we did not exploit this knowledge for training our multi-domain methods, but instead allowed them to discover the relations autonomously.  

\subsection{Dialog state tracking}
The first step for training our RL agent is to train a DST model to use for the RL module's observations. For this, we used the original SimDial data set, which consists of 2,000 training/validation dialogs and 500 test dialogs per domain. We held out 400 dialogs for validation from each domain, and trained the model in a batch MTL fashion. Additional details on the data set are presented in Table~\ref{tab:SimDial}.

\begin{table}[t]
    \centering
    \caption{Summary of the SimDial data set~\citep{zhao2018zeroshot}. User slots are informed by the user and system slots by the agent. Slot values are the number of values that each user slot can take.}
    \label{tab:SimDial}
    \begin{tabular}{c|c|c|c|c}
         Domain	& \# User slots	& \# Slot values	& \# System slots	& Avg. length\\
        \hline
        \hline
Bus	& 3	& 13, 13, 30	& 2	& 10\\
\hline
Movie	& 3	& 15, 7, 10	& 3	& 11\\
\hline
\begin{tabular}[c]{@{}c@{}}Rest\\Rest\_slot\\Rest\_style\end{tabular}	& 2	& 11, 13	& 3 	& 8\\
\hline
Weather	& 2	& 11, 7	& 2	& 8
    \end{tabular}
\end{table}

In order to train the DST, we used the architecture from Figure~\ref{fig:MultiDomainDST}. We used word embeddings of 400 neurons, utterance embeddings of 300 neurons with bidirectional long short-term memory (bi-LSTM), dialog embeddings of 200 neurons with regular LSTM, and a dense layer of 200 neurons. All hidden layers were transformed with rectified linear (ReLU) activation. Results using this architecture are presented in Table~\ref{tab:SimDialDST}, which shows the joint test accuracy across all slots on each domain.

\begin{table}[t]
    \centering
    \caption{Slot tracking accuracy of the MTL DST on each domain. Joint accuracy is measured by only considering an instance as correctly tracked if all slot values were predicted correctly. }
    \label{tab:SimDialDST}
    \begin{tabular}{c|c|c|c|c|c|c}
                 &  Bus   &  Movie  & Rest  & Rest slot & Rest style & Weather\\
        \hline
        Joint accuracy & 69.58\% & 79.43\% & 86.66\% & 85.26\%     & 88.08\% & 87.34\%
    \end{tabular}
\end{table}

\subsection{Multi-domain RL}
Once the model for DST was trained, we finalized our construction of the MDP. As mentioned in Section~\ref{sec:MultiDomainArchitecture}, the observations came from the hidden state of the dialog-level LSTM in the DST module. The actions were \textit{request}, \textit{confirm}, and \textit{inform}, for each relevant slot in each domain. Note that, even though the three restaurant domains shared the slots (albeit with different specific values in the restaurant-slot domain), we did not explicitly encode this in our model, letting each domain have a separate action prediction head. In practice, this means that our model had to autonomously discover these relations across the domains. As mentioned in Section~\ref{sec:DialogPolicyLearning}, the transition function is determined by the user. We leveraged the simulator used to construct the SimDial data set to simulate users. We imposed a maximum length of $15$ turns per dialog. Finally, the reward function was chosen to be $-1$ for every step taken, to encourage the agent to quickly solve the user's requests, and $+30$ at the final step for satisfying the user's overall goal. An example user goal is to find a Chinese restaurant in New York and get the price range of the restaurant. Note that the system acts are transmitted directly to the user simulator, so that the NLG does not play a role in determining the success of the task.

As a first baseline, we used the policy that generated the SimDial corpus, without implicit confirmation actions, as these were only executed in combination with another action (e.g., requesting an additional value) and RL agents are not capable of handling multiple simultaneous actions. Note that this is a powerful baseline, as 1) it receives ground-truth user actions to use for state tracking, and 2) the user simulator was designed specifically to work in conjunction with this policy.  For a second baseline, we trained separate TRPO agents with the same architecture from Figure~\ref{fig:MultiDomainRL} for each domain. The  action embedding layer was fixed to 100 neurons. We ran a grid search to select the optimal hyper-parameters for TRPO in terms of success rate after 10,000 dialogs of training. In particular, we optimized over the number of dialogs per iteration, from $\{10, 20, 50, 100, 200, 500, 1000\}$, and the maximum KL divergence between consecutive policies, from $\{0.001, 0.003, 0.005, 0.01, 0.03, 0.05, 0.1, 0.3, 0.5\}$. We kept this architecture and hyper-parameters fixed for our MTL and TL methods, forgoing potential additional benefits of tuning the hyper-parameters specifically for them in favor of simplicity. MTL training was done on all domains simultaneously, while TL training was done in a leave-one-out fashion, using each domain in turn as the target domain. 
\begin{table}[t]
    \centering
    \caption{Dialog policy success rate on the SimDial simulator. Both multi-domain variants learn with considerably fewer interactions than single-domain TRPO, and achieve consistently better performance after convergence. Standard errors reported after the $\pm$}
    \label{tab:SimDialSuccess}
    \begin{tabular}{l|l|r@{}l|r@{}l|r@{}l}
         Domain      & Algorithm     & Success& \,\,@ 2k      & Dialogs > Ru&le-based   & Success & \,\,@10k\\
\hline
\hline
\multirow{4}{*}{Bus}     & Rule-based     & ---&   & ---&   & $52.8\pm$&$0.5\%$\\
     & TRPO  & $56\pm$&$2\%$     & $1680\pm$&$103$      & $82.8\pm$&$0.6$\\
     & TRPO MTL      & $\bm{63\pm}$&$\bm{2\%}$    & $\bm{1290\pm}$&$\bm{84}$      & $\bm{84.1\pm}$&$\bm{0.7\%}$\\
     & TRPO TL   & $\bm{62\pm}$&$\bm{2\%}$    & $\bm{1270\pm}$&$\bm{76}$      & $\bm{83.9\pm}$&$\bm{0.5\%}$\\
\hline
\multirow{4}{*}{Movie}   & Rule-based     & ---&   & ---&   & $49.9\pm$&$0.5\%$\\
     & TRPO  & $42\pm$&$2\%$     & $2520\pm$&$154$      & $83.1\pm$&$0.9\%$\\
     & TRPO MTL      & $\bm{59\pm}$&$\bm{2\%}$    & $\bm{1520\pm}$&$\bm{75}$      & $\bm{87.6\pm}$&$\bm{1.1\%}$\\
     & TRPO TL   & $\bm{58\pm}$&$\bm{2\%}$    & $\bm{1570\pm}$&$\bm{96}$      & $84.2\pm$&$0.7\%$\\
\hline
\multirow{4}{*}{Rest slot}   & Rule-based     & ---&   & ---&   & $69.3\pm$&$0.5\%$\\
     & TRPO  & $57\pm$&$2\%$     & $2560\pm$&$175$      & $90.8\pm$&$0.6\%$\\
     & TRPO MTL      & $\bm{70\pm}$&$\bm{2\%}$    & $\bm{1860\pm}$&$\bm{89}$      & $\bm{93.6\pm}$&$\bm{0.4\%}$\\
     & TRPO TL   & $\bm{71\pm}$&$\bm{2\%}$    & $\bm{1710\pm}$&$\bm{97}$      & $92.5\pm$&$0.3\%$\\
\hline
\multirow{4}{*}{Rest}    & Rule-based     & ---&   & ---&   & $68.8\pm$&$0.5$\\
     & TRPO      & $50\pm$&$2\%$     & $3120\pm$&$168$      & $89.1\pm$&$0.6\%$\\
     & TRPO MTL      & $\bm{67\pm}$&$\bm{2\%}$    & $\bm{1970\pm}$&$\bm{120}$     & $\bm{92.9\pm}$&$\bm{0.5\%}$\\
     & TRPO TL   & $\bm{65\pm}$&$\bm{2\%}$    & $\bm{2020\pm}$&$\bm{82}$      & $90.6\pm$&$0.7$\\
\hline
\multirow{4}{*}{Rest style}      & Rule-based     & ---&   & ---&   & $68.3\pm$&$0.4\%$\\
     & TRPO  & $51\pm$&$2\%$     & $3130\pm$&$183$      & $88.5\pm$&$0.6\%$\\
     & TRPO MTL      & $\bm{69\pm}$&$\bm{2\%}$    & $\bm{1710\pm}$&$\bm{91}$      & $\bm{92.4\pm}$&$\bm{0.5\%}$\\
     & TRPO TL   & $\bm{65\pm}$&$\bm{3\%}$    & $\bm{1910\pm}$&$\bm{120}$     & $90.5\pm$&$0.6\%$\\
\hline
\multirow{4}{*}{Weather}     & Rule-based     & ---&   & ---&   & $77.6\pm$&$0.4$\\
     & TRPO      & $78\pm$&$1\%$     & $1840\pm$&$94$   & $\bm{95.0\pm}$&$\bm{0.4\%}$\\
     & TRPO MTL      & $\bm{85\pm}$&$\bm{1\%}$    & $\bm{1170\pm}$&$\bm{79}$      & $\bm{95.6\pm}$&$\bm{0.3\%}$\\
     & TRPO TL   & $\bm{83\pm}$&$\bm{3\%}$    & $\bm{1240\pm}$&$\bm{124}$     & $\bm{95.2\pm}$&$\bm{0.4\%}$\\
\hline
\hline
\multirow{4}{*}{Average}     & Rule-based     & ---&   & ---&   & $64.5\pm$&$0.2\%$\\
     & TRPO  & $56\pm$&$1\%$     & $2475\pm$&$91$   & $88.2\pm$&$0.2\%$\\
     & TRPO MTL      & $\bm{69\pm}$&$\bm{1\%}$    & $\bm{1587\pm}$&$\bm{47}$      & $\bm{91.0\pm}$&$\bm{0.3\%}$\\
     & TRPO TL   & $\bm{68\pm}$&$\bm{1\%}$    & $\bm{1620\pm}$&$\bm{39}$      & $89.5\pm$&$0.3\%$\\
    \end{tabular}
\end{table}
\begin{table}[t]
    \centering
    \caption{Average length of learned dialogs on the SimDial simulator. The multi-domain methods not only have a higher success rate, but are also faster in solving the user's queries than single-domain TRPO. Standard errors reported after the $\pm$.}
    \label{tab:SimDialLength}
    \begin{tabular}{l|l|r@{}l|r@{}l|r@{}l}
         Domain      & Algorithm    & Length \,\,& @ 2k       & Dialogs < R &ule-based     & Length @&\,\, 10k\\
\hline
\hline
\multirow{4}{*}{Bus}      & Rule-based    & ---&     & ---&     & $12.69\pm$&$0.02$\\
      & TRPO    & $11.9\pm$&$0.2$   & $1010\pm$&$68$    & $9.30\pm$&$0.08$\\
      & TRPO MTL     & $\bm{11.6\pm}$&$\bm{0.1}$   & $\bm{910\pm}$&$\bm{90}$    & $\bm{9.03\pm}$&$\bm{0.10}$\\
      & TRPO TL      & $\bm{11.5\pm}$&$\bm{0.1}$   & $\bm{900\pm}$&$\bm{55}$    & $\bm{9.09\pm}$&$\bm{0.07}$\\
\hline
\multirow{4}{*}{Movie}    & Rule-based    & ---&     & ---&     & $12.88\pm$&$0.04$\\
      & TRPO    & $12.8\pm$&$0.1$   & $1790\pm$&$134$   & $9.81\pm$&$0.08$\\
      & TRPO MTL     & $\bm{11.9\pm}$&$\bm{0.2}$      & $\bm{1200\pm}$&$\bm{49}$    & $\bm{9.31\pm}$&$\bm{0.13}$\\
      & TRPO TL      & $\bm{11.9\pm}$&$\bm{0.1}$   & $\bm{1210\pm}$&$\bm{94}$    & $9.57\pm$&$0.06$\\
\hline
\multirow{4}{*}{Rest slot}     & Rule-based    & ---&     & ---&     & $11.03\pm$&$0.03$\\
      & TRPO    & $11.6\pm$&$0.2$   & $2100\pm$&$144$   & $8.12\pm$&$0.09$\\
      & TRPO MTL     & $\bm{10.6\pm}$&$\bm{0.2}$   & $\bm{1520\pm}$&$\bm{102}$   & $\bm{7.70\pm}$&$\bm{0.07}$\\
      & TRPO TL      & $\bm{10.5\pm}$&$\bm{0.2}$   & $\bm{1550\pm}$&$\bm{72}$    & $\bm{7.79\pm}$&$\bm{0.06}$\\
\hline
\multirow{4}{*}{Rest}      & Rule-based    & ---&     & ---&     & $10.98\pm$&$0.04$\\
      & TRPO    & $12.1\pm$&$0.2$   & $2460\pm$&$141$   & $8.50\pm$&$0.06$\\
      & TRPO MTL     & $\bm{10.9\pm}$&$\bm{0.2}$      & $\bm{1720\pm}$&$\bm{91}$    & $\bm{7.97\pm}$&$\bm{0.07}$\\
      & TRPO TL      & $\bm{10.8\pm}$&$\bm{0.2}$   & $\bm{1810\pm}$&$\bm{108}$   & $8.14\pm$&$0.08$\\
\hline
\multirow{4}{*}{Rest style}    & Rule-based    & ---&     & ---&     & $11.01\pm$&$0.03$\\
      & TRPO    & $12.1\pm$&$0.2$   & $2680\pm$&$155$   & $8.41\pm$&$0.05$\\
      & TRPO MTL     & $\bm{10.5\pm}$&$\bm{0.2}$   & $\bm{1510\pm}$&$\bm{102}$   & $\bm{8.00\pm}$&$\bm{0.05}$\\
      & TRPO TL      & $\bm{10.8\pm}$&$\bm{0.2}$   & $\bm{1750\pm}$&$\bm{162}$   & $8.21\pm$&$0.09$\\
\hline
\multirow{4}{*}{Weather}       & Rule-based    & ---&     & ---&     & $10.26\pm$&$0.03$\\
      & TRPO    & $9.9\pm$&$0.2$    & $1490\pm$&$100$   & $6.62\pm$&$0.07$\\
      & TRPO MTL     & $\bm{8.7\pm}$&$\bm{0.1}$    & $\bm{910\pm}$&$\bm{41}$    & $\bm{6.46\pm}$&$\bm{0.07}$\\
      & TRPO TL      & $\bm{9.0\pm}$&$\bm{0.3}$    & $\bm{1010\pm}$&$\bm{75}$    & $6.71\pm$&$0.06$\\
\hline
\hline
\multirow{4}{*}{Average}       & Rule-based    & ---&     & ---&     & $11.48\pm$&$0.01$\\
      & TRPO    & $11.7\pm$&$0.1$   & $1922\pm$&$76$    & $8.46\pm$&$0.03$\\
      & TRPO MTL     & $\bm{10.7\pm}$&$\bm{0.1}$      & $\bm{1295\pm}$&$\bm{42}$    & $\bm{8.08\pm}$&$\bm{0.04}$\\
      & TRPO TL      & $\bm{10.8\pm}$&$\bm{0.1}$      & $\bm{1372\pm}$&$\bm{50}$    & $8.25\pm$&$0.03$\\
    \end{tabular}
\end{table}

Table~\ref{tab:SimDialSuccess} contains comparisons of the success rates of policies trained via MTL and TL against our baselines, averaged over 10 random seeds. The first column shows the performance of all methods after only 2,000 dialogs with users in each domain, where our MTL and TL methods substantially outperformed the single-domain baseline by an average of $23\%$ and $21\%$, respectively. The second column shows the number of dialogs that were needed, on average, to outperform the rule-based baseline. This is a crucial metric, as the rule-based policy is quite a strong baseline given that it was designed by the creators of the data set in conjunction with the user simulator. In both multi-domain cases, we saw a reduction of approximately $35\%$ in the number of dialogs needed to outperform the baseline. Finally, we provide the performance after 10,000 dialogs per domain, at which time all methods had already converged. Even after the single-domain method was allowed to see sufficient dialogs to converge, the MTL and TL methods outperformed it by $3.2\%$ and $1.5\%$, respectively. The results on Table~\ref{tab:SimDialLength} show that the same analysis also applies to the length of the dialogs generated by our policy, which was consistently more efficient at solving the user's queries than the baselines.

\section{Conclusions}
We proposed the use of domain-agnostic action embeddings for accelerating the learning of task-oriented dialog policies via RL. Intuitively, these action embeddings autonomously discover how policies for acting in different domains relate to each other. We showed how our approach is capable of more quickly and more effectively learning to solve the problem than learning separate domain-specific policies in our evaluation on the SimDial data set.

In this work, we explored two different modalities for training the multi-domain policies: batch multi-task learning and transfer learning. In future work, we plan to explore how to train these policies in a \textit{continual learning} setting, where domains are encountered by the system sequentially. Another interesting direction would be to extend current meta-learning algorithms to the multi-domain setting, where the output space is not shared across different domains, as is the case in dialog systems. Doing so could potentially further accelerate the learning in the transfer setting explored here. One additional line of work would be exploring how much additional benefit could be obtained by pre-training the policies via supervised learning.

\bibliographystyle{plainnat}
\bibliography{MultiDomainRL}

\begin{thebibliography}{32}
\providecommand{\natexlab}[1]{#1}
\providecommand{\url}[1]{\texttt{#1}}
\expandafter\ifx\csname urlstyle\endcsname\relax
  \providecommand{\doi}[1]{doi: #1}\else
  \providecommand{\doi}{doi: \begingroup \urlstyle{rm}\Url}\fi

\bibitem[Bordes et~al.(2017)Bordes, Boureau, and Weston]{bordes2017learning}
Antoine Bordes, Y{-}Lan Boureau, and Jason Weston.
\newblock Learning end-to-end goal-oriented dialog.
\newblock In \emph{Proceedings of the 5th International Conference on Learning
  Representations (ICLR-17)}, 2017.
\newblock URL \url{https://openreview.net/forum?id=S1Bb3D5gg}.

\bibitem[Budzianowski et~al.(2017)Budzianowski, Ultes, Su, Mrk{\v{s}}i{\'c},
  Wen, Casanueva, Rojas-Barahona, and Ga{\v{s}}i{\'c}]{budzianowski2017sub}
Pawe{\l} Budzianowski, Stefan Ultes, Pei-Hao Su, Nikola Mrk{\v{s}}i{\'c},
  Tsung-Hsien Wen, I{\~n}igo Casanueva, Lina~M. Rojas-Barahona, and Milica
  Ga{\v{s}}i{\'c}.
\newblock Sub-domain modelling for dialogue management with hierarchical
  reinforcement learning.
\newblock In \emph{Proceedings of the 18th Annual Meeting of the Special
  Interest Group on Discourse and Dialogue (SIGDIAL-17)}. Association for
  Computational Linguistics, 2017.

\bibitem[Chen et~al.(2017)Chen, Liu, Yin, and Tang]{chen2017survey}
Hongshen Chen, Xiaorui Liu, Dawei Yin, and Jiliang Tang.
\newblock A survey on dialogue systems: Recent advances and new frontiers.
\newblock \emph{ACM Special Interest Group on Knowledge Discovery and Data
  Mining Explorations Newsletter (SIGKDD-17)}, 19\penalty0 (2):\penalty0
  25--35, 2017.

\bibitem[Chen et~al.(2018)Chen, Chang, Chen, Tan, Ga{\v{s}}i{\'c}, and
  Yu]{chen2018policy}
Lu~Chen, Cheng Chang, Zhi Chen, Bowen Tan, Milica Ga{\v{s}}i{\'c}, and Kai Yu.
\newblock Policy adaptation for deep reinforcement learning-based dialogue
  management.
\newblock In \emph{2018 IEEE International Conference on Acoustics, Speech and
  Signal Processing (ICASSP-18)}, pages 6074--6078. IEEE, 2018.

\bibitem[Cuay{\'a}huitl et~al.(2016)Cuay{\'a}huitl, Yu, Williamson, and
  Carse]{cuayhuitl2016deep}
Heriberto Cuay{\'a}huitl, Seunghak Yu, Ashley Williamson, and Jacob Carse.
\newblock Deep reinforcement learning for multi-domain dialogue systems.
\newblock In \emph{3rd Deep Reinforcement Learning Workshop at Neural
  Information Processing Systems 30 (NeurIPS-DeepRL-16)}, 2016.

\bibitem[Cuay{\'a}huitl et~al.(2017)Cuay{\'a}huitl, Yu, Williamson, and
  Carse]{cuayahuitl2017scaling}
Heriberto Cuay{\'a}huitl, Seunghak Yu, Ashley Williamson, and Jacob Carse.
\newblock Scaling up deep reinforcement learning for multi-domain dialogue
  systems.
\newblock In \emph{2017 International Joint Conference on Neural Networks
  (IJCNN-17)}, pages 3339--3346. IEEE, 2017.

\bibitem[Dulac-Arnold et~al.(2015)Dulac-Arnold, Evans, van Hasselt, Sunehag,
  Lillicrap, Hunt, Mann, Weber, Degris, and Coppin]{dulac2015deep}
Gabriel Dulac-Arnold, Richard Evans, Hado van Hasselt, Peter Sunehag, Timothy
  Lillicrap, Jonathan Hunt, Timothy Mann, Theophane Weber, Thomas Degris, and
  Ben Coppin.
\newblock Deep reinforcement learning in large discrete action spaces.
\newblock \emph{arXiv preprint arXiv:1512.07679}, 2015.

\bibitem[Fazel-Zarandi et~al.(2017)Fazel-Zarandi, Li, Cao, Casale, Henderson,
  Whitney, and Geramifard]{fazel2017learning}
Maryam Fazel-Zarandi, Shang-Wen Li, Jin Cao, Jared Casale, Peter Henderson,
  David Whitney, and Alborz Geramifard.
\newblock Learning robust dialog policies in noisy environments.
\newblock In \emph{1st Workshop on Conversational {AI} at Neural Information
  Processing Systems 31 (NeurIPS-ConvAI-17)}, 2017.

\bibitem[Gao et~al.(2019)Gao, Galley, and Li]{Gao2019Neural}
Jianfeng Gao, Michel Galley, and Lihong Li.
\newblock Neural approaches to conversational ai.
\newblock \emph{Foundations and Trends® in Information Retrieval}, 13\penalty0
  (2-3):\penalty0 127--298, 2019.
\newblock ISSN 1554-0669.
\newblock \doi{10.1561/1500000074}.
\newblock URL \url{http://dx.doi.org/10.1561/1500000074}.

\bibitem[Ga{\v{s}}i{\'c} et~al.(2013)Ga{\v{s}}i{\'c}, Breslin, Henderson, Kim,
  Szummer, Thomson, Tsiakoulis, and Young]{gasic2013pomdp}
Milica Ga{\v{s}}i{\'c}, Catherine Breslin, Matthew Henderson, Dongho Kim,
  Martin Szummer, Blaise Thomson, Pirros Tsiakoulis, and Steve Young.
\newblock {POMDP}-based dialogue manager adaptation to extended domains.
\newblock In \emph{Proceedings of the 14th Annual Meeting of the Special
  Interest Group on Discourse and Dialogue (SIGDIAL-13)}, pages 214--222, Metz,
  France, 2013. Association for Computational Linguistics.
\newblock URL \url{https://www.aclweb.org/anthology/W13-4035}.

\bibitem[Ga{\v{s}}i{\'c} et~al.(2017)Ga{\v{s}}i{\'c}, Mrk{\v{s}}i{\'c},
  Rojas-Barahona, Su, Ultes, Vandyke, Wen, and Young]{gavsic2017dialogue}
Milica Ga{\v{s}}i{\'c}, Nikola Mrk{\v{s}}i{\'c}, Lina~M Rojas-Barahona, Pei-Hao
  Su, Stefan Ultes, David Vandyke, Tsung-Hsien Wen, and Steve Young.
\newblock Dialogue manager domain adaptation using {G}aussian process
  reinforcement learning.
\newblock \emph{Computer Speech \& Language}, 45:\penalty0 552--569, 2017.

\bibitem[He et~al.(2016)He, Chen, He, Gao, Li, Deng, and Ostendorf]{he2016deep}
Ji~He, Jianshu Chen, Xiaodong He, Jianfeng Gao, Lihong Li, Li~Deng, and Mari
  Ostendorf.
\newblock Deep reinforcement learning with a natural language action space.
\newblock In \emph{Proceedings of the 54th Annual Meeting of the Association
  for Computational Linguistics (Volume 1: Long Papers)}, pages 1621--1630,
  2016.

\bibitem[Kim et~al.(2014)Kim, Henderson, Ga{\v{s}}i{\'c}, Tsiakoulis, and
  Young]{kim2014use}
Dongho Kim, Matthew Henderson, Milica Ga{\v{s}}i{\'c}, Pirros Tsiakoulis, and
  Steve Young.
\newblock The use of discriminative belief tracking in {POMDP}-based dialogue
  systems.
\newblock In \emph{2014 IEEE Spoken Language Technology Workshop (SLT-14)},
  pages 354--359. IEEE, 2014.

\bibitem[{Lee}(2017)]{lee2017toward}
Sungjin {Lee}.
\newblock {Toward Continual Learning for Conversational Agents}.
\newblock \emph{arXiv e-prints}, art. arXiv:1712.09943, Dec 2017.

\bibitem[Lipton et~al.(2018)Lipton, Li, Gao, Li, Ahmed, and
  Deng]{lipton2018bbqnetworks}
Zachary Lipton, Xiujun Li, Jianfeng Gao, Lihong Li, Faisal Ahmed, and Li~Deng.
\newblock {BBQ-}networks: Efficient exploration in deep reinforcement learning
  for task-oriented dialogue systems.
\newblock In \emph{Proceedings of the 32nd AAAI Conference on Artificial
  Intelligence (AAAI-18)}, 2018.
\newblock URL
  \url{https://aaai.org/ocs/index.php/AAAI/AAAI18/paper/view/16189}.

\bibitem[Liu and Lane(2017)]{liu2017multidomain}
Bing Liu and Ian Lane.
\newblock Multi-domain adversarial learning for slot filling in spoken language
  understanding.
\newblock In \emph{1st Workshop on Conversational {AI} at Neural Information
  Processing Systems 31 (NeurIPS-ConvAI-17)}, 2017.

\bibitem[Liu et~al.(2017)Liu, Tur, Hakkani-Tur, Shah, and Heck]{liu2017endto}
Bing Liu, Gokhan Tur, Dilek Hakkani-Tur, Pararth Shah, and Larry Heck.
\newblock End-to-end optimization of task-oriented dialogue model with deep
  reinforcement learning.
\newblock In \emph{1st Workshop on Conversational {AI} at Neural Information
  Processing Systems 31 (NeurIPS-ConvAI-17)}, 2017.

\bibitem[Mnih et~al.(2015)Mnih, Kavukcuoglu, Silver, Rusu, Veness, Bellemare,
  Graves, Riedmiller, Fidjeland, Ostrovski, et~al.]{mnih2015human}
Volodymyr Mnih, Koray Kavukcuoglu, David Silver, Andrei~A Rusu, Joel Veness,
  Marc~G Bellemare, Alex Graves, Martin Riedmiller, Andreas~K Fidjeland, Georg
  Ostrovski, et~al.
\newblock Human-level control through deep reinforcement learning.
\newblock \emph{Nature}, 518\penalty0 (7540):\penalty0 529, 2015.

\bibitem[Peng et~al.(2017)Peng, Li, Li, Gao, Celikyilmaz, Lee, and
  Wong]{peng2017composite}
Baolin Peng, Xiujun Li, Lihong Li, Jianfeng Gao, Asli Celikyilmaz, Sungjin Lee,
  and Kam-Fai Wong.
\newblock Composite task-completion dialogue policy learning via hierarchical
  deep reinforcement learning.
\newblock In \emph{Proceedings of the 2017 Conference on Empirical Methods in
  Natural Language Processing (EMNLP-17)}, pages 2231--2240, Copenhagen,
  Denmark, 2017. Association for Computational Linguistics.
\newblock \doi{10.18653/v1/D17-1237}.
\newblock URL \url{https://www.aclweb.org/anthology/D17-1237}.

\bibitem[Qian and Yu(2019)]{qian2019domain}
Kun Qian and Zhou Yu.
\newblock Domain adaptive dialog generation via meta learning.
\newblock In \emph{Proceedings of the 57th Annual Meeting of the Association
  for Computational Linguistics (ACL-19)}, pages 2639--2649, Florence, Italy,
  2019. Association for Computational Linguistics.
\newblock \doi{10.18653/v1/P19-1253}.
\newblock URL \url{https://www.aclweb.org/anthology/P19-1253}.

\bibitem[Rastogi et~al.(2017)Rastogi, Hakkani-T{\"u}r, and
  Heck]{rastogi2017scalable}
Abhinav Rastogi, Dilek Hakkani-T{\"u}r, and Larry Heck.
\newblock Scalable multi-domain dialogue state tracking.
\newblock In \emph{2017 IEEE Automatic Speech Recognition and Understanding
  Workshop (ASRU-17)}, pages 561--568. IEEE, 2017.

\bibitem[Schulman et~al.(2015)Schulman, Levine, Abbeel, Jordan, and
  Moritz]{schulman2015trust}
John Schulman, Sergey Levine, Pieter Abbeel, Michael Jordan, and Philipp
  Moritz.
\newblock Trust region policy optimization.
\newblock In Francis Bach and David Blei, editors, \emph{Proceedings of the
  32nd International Conference on Machine Learning (ICML-15)}, volume~37 of
  \emph{Proceedings of Machine Learning Research}, pages 1889--1897, Lille,
  France, 07--09 Jul 2015. PMLR.

\bibitem[Shah et~al.(2019)Shah, Gupta, Fayazi, and Hakkani-Tur]{shah2019robust}
Darsh Shah, Raghav Gupta, Amir Fayazi, and Dilek Hakkani-Tur.
\newblock Robust zero-shot cross-domain slot filling with example values.
\newblock In \emph{Proceedings of the 57th Annual Meeting of the Association
  for Computational Linguistics (ACL-19)}, pages 5484--5490, Florence, Italy,
  2019. Association for Computational Linguistics.
\newblock \doi{10.18653/v1/P19-1547}.
\newblock URL \url{https://www.aclweb.org/anthology/P19-1547}.

\bibitem[Shu et~al.(2018)Shu, Molino, Namazifar, Liu, Xu, Zheng, and
  Tur]{shu2018incorporating}
Lei Shu, Piero Molino, Mahdi Namazifar, Bing Liu, Hu~Xu, Huaixiu Zheng, and
  Gokhan Tur.
\newblock Incorporating the structure of the belief state in end-to-end
  task-oriented dialogue systems.
\newblock In \emph{2nd Workshop on Conversational {AI} at Neural Information
  Processing Systems 32 (NeurIPS-ConvAI-18)}, 2018.

\bibitem[Tang et~al.(2018)Tang, Li, Gao, Wang, Li, and Jebara]{tang2018subgoal}
Da~Tang, Xiujun Li, Jianfeng Gao, Chong Wang, Lihong Li, and Tony Jebara.
\newblock Subgoal discovery for hierarchical dialogue policy learning.
\newblock In \emph{Proceedings of the 2018 Conference on Empirical Methods in
  Natural Language Processing (EMNLP-18)}, October-November 2018.

\bibitem[Vlasov et~al.(2018)Vlasov, Drissner-Schmid, and
  Nichol]{vlasov2018fewshot}
Vladimir Vlasov, Akela Drissner-Schmid, and Alan Nichol.
\newblock Few-shot generalization across dialogue tasks.
\newblock In \emph{2nd Workshop on Conversational {AI} Neural Information
  Processing Systems 32 (NeurIPS-ConvAI-18)}, 2018.

\bibitem[Wang et~al.(2014)Wang, Chen, Wang, Tian, Wu, and Wang]{wang2014policy}
Zhuoran Wang, Hongliang Chen, Guanchun Wang, Hao Tian, Hua Wu, and Haifeng
  Wang.
\newblock Policy learning for domain selection in an extensible multi-domain
  spoken dialogue system.
\newblock In \emph{Proceedings of the 2014 Conference on Empirical Methods in
  Natural Language Processing (EMNLP-14)}, pages 57--67, Doha, Qatar, 2014.
  Association for Computational Linguistics.
\newblock \doi{10.3115/v1/D14-1007}.
\newblock URL \url{https://www.aclweb.org/anthology/D14-1007}.

\bibitem[Wang et~al.(2015)Wang, Wen, Su, and Stylianou]{wang2015learning}
Zhuoran Wang, Tsung-Hsien Wen, Pei-Hao Su, and Yannis Stylianou.
\newblock Learning domain-independent dialogue policies via ontology
  parameterisation.
\newblock In \emph{Proceedings of the 16th Annual Meeting of the Special
  Interest Group on Discourse and Dialogue (SIGDIAL-15)}, pages 412--416,
  Prague, Czech Republic, September 2015. Association for Computational
  Linguistics.
\newblock \doi{10.18653/v1/W15-4654}.
\newblock URL \url{https://www.aclweb.org/anthology/W15-4654}.

\bibitem[Williams et~al.(2017)Williams, Asadi, and Zweig]{williams2017hybrid}
Jason~D. Williams, Kavosh Asadi, and Geoffrey Zweig.
\newblock Hybrid code networks: practical and efficient end-to-end dialog
  control with supervised and reinforcement learning.
\newblock In \emph{Proceedings of the 55th Annual Meeting of the Association
  for Computational Linguistics (ACL-17)}, pages 665--677, Vancouver, Canada,
  2017. Association for Computational Linguistics.
\newblock \doi{10.18653/v1/P17-1062}.
\newblock URL \url{https://www.aclweb.org/anthology/P17-1062}.

\bibitem[Zhao and Eskenazi(2016)]{zhao2016towards}
Tiancheng Zhao and Maxine Eskenazi.
\newblock Towards end-to-end learning for dialog state tracking and management
  using deep reinforcement learning.
\newblock In \emph{Proceedings of the 17th Annual Meeting of the Special
  Interest Group on Discourse and Dialogue (SIGDIAL-16)}, pages 1--10, Los
  Angeles, September 2016. Association for Computational Linguistics.
\newblock URL \url{http://www.aclweb.org/anthology/W16-3601}.

\bibitem[Zhao and Eskenazi(2018)]{zhao2018zeroshot}
Tiancheng Zhao and Maxine Eskenazi.
\newblock Zero-shot dialog generation with cross-domain latent actions.
\newblock In \emph{Proceedings of the 19th Annual Meeting of the Special
  Interest Group on Discourse and Dialogue (SIGDIAL-18)}, pages 1--10,
  Melbourne, Australia, July 2018. Association for Computational Linguistics.

\bibitem[Zhou et~al.(2017)Zhou, Small, Rokhlenko, and Elkan]{zhou2017endto}
Li~Zhou, Kevin Small, Oleg Rokhlenko, and Charles Elkan.
\newblock End-to-end offline goal-oriented dialog policy learning via policy
  gradient.
\newblock In \emph{1st Workshop on Conversational {AI} at Neural Information
  Processing Systems 31 (NeurIPS-ConvAI-17)}, 2017.

\end{thebibliography}

\end{document}